\documentclass[runningheads]{llncs}

\usepackage{eccv}

\usepackage{eccvabbrv}

\usepackage{graphicx}
\usepackage{booktabs}

\usepackage[accsupp]{axessibility}  % Improves PDF readability for those with disabilities.

\usepackage{multirow}
\newcommand{\best}[1]{\textbf{#1}}
\newcommand{\second}[1]{\underline{#1}}

\usepackage{makecell} 
\usepackage{hyperref}

\usepackage{orcidlink}

\usepackage{svg}
\usepackage{amsmath}  
\begin{document}

\title{FUMO: Prior-Modulated Diffusion for Single Image Reflection Removal} 

\titlerunning{FUMO}

\author{Telang Xu\inst{1}\textsuperscript{*}\orcidlink{0009-0004-8921-4124} \and
Chaoyang Zhang\inst{2,3}\textsuperscript{*}\orcidlink{0009-0007-5341-2657} \and
Guangtao Zhai\inst{1}\orcidlink{0000-0001-8165-9322} \and \\
Xiaohong Liu\inst{1,2}\textsuperscript{$\dagger$}\orcidlink{0000-0001-6377-4730}}

\authorrunning{T.~Xu et al.}

\institute{Shanghai Jiao Tong University, Shanghai, China \and
Shanghai Innovation Institute, Shanghai, China  \and
Xi'an Jiaotong University, Xi'an, China \\
\email{\{luciousdesmon, xiaohongliu, zhaiguangtao\}@sjtu.edu.cn \{zcyxjtu65\}@gmail.com}\\
\textsuperscript{*}Equal Contribution \textsuperscript{$\dagger$}Corresponding Author
}

\maketitle

\begin{abstract}
Single image reflection removal (SIRR) is challenging in real scenes, where reflection strength varies spatially and reflection patterns are tightly entangled with transmission structures. This paper presents a diffusion model with prior modulation framework (FUMO) that introduces explicit priors for spatially adaptive conditioning and structurally faithful restoration. Two priors are extracted directly from the mixed image, an intensity prior that estimates spatial reflection severity and a high-frequency prior that captures detail-sensitive responses via multi-scale residual aggregation. We propose a coarse-to-fine training paradigm. In the first stage, these cues are combined to gate the conditional residual injections, focusing the conditioning on regions that are both reflection-dominant and structure-sensitive. In the second stage, a fine-grained refinement network corrects local misalignment and sharpens fine details in the image space. Experiments conducted on both standard benchmarks and challenging images in the wild demonstrate competitive quantitative results and consistently improved perceptual quality. The code is released at \href{https://github.com/Lucious-Desmon/FUMO}{https://github.com/Lucious-Desmon/FUMO}.
\end{abstract}

\section{Introduction}
\label{sec:intro}

\begin{figure}[t]
    \centering
    \includegraphics[width=\linewidth]{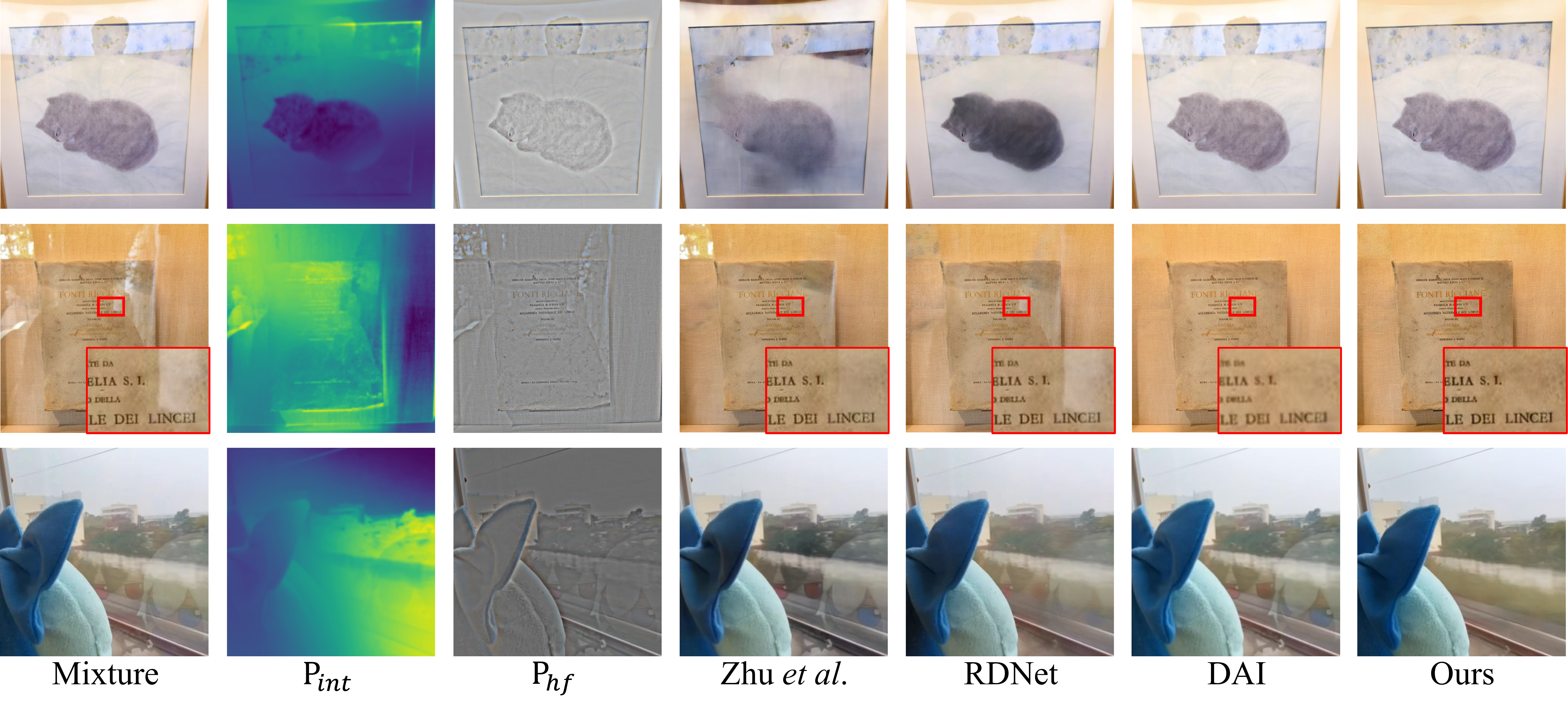}
    \caption{Failure-mode visualization on in-the-wild reflection mixtures. Three representative real-world mixed images are shown together with two priors. Qualitative comparisons with representative SOTA methods illustrate common challenges in the wild, including incomplete reflection suppression, color inconsistency, and loss of fine details. Red rectangles highlight regions for closer inspection.}
    \label{fig:intro_problem}
\end{figure}

Images captured through glass or other transparent media often contain unwanted reflections that obscure the underlying scene~\cite{Yang_2025_CVPR,wan2017benchmarking}. Such artifacts degrade visual quality and hinder downstream vision applications~\cite{liu2020reflection,wan2021face}. Reflection removal aims to suppress reflections and recover a clear transmission image from a mixture input. The task in real-world settings remains challenging due to severe layer entanglement and large spatial variations in reflection appearance ~\cite{dong2021location,hu2023single,zhu2024revisiting}.

Methods~\cite{Simon_2015_CVPR,yao2025polarfree,yang2016robust,szeliski2000layer,lei2020polarized,agrawal2005removing,han2017reflection} leveraging multiple images depend on specialized acquisition and are less applicable to casual photography or internet images. In the single-image task, traditional optimization methods~\cite{Arvanitopoulos_2017_CVPR,lei2021robust,zheng2021single,xue2015computational,levin2007user,shih2015reflection,zhong2024language} impose hand-crafted priors to decompose layers, yet they are often brittle when assumptions are violated and computationally expensive. Learning-based approaches~\cite{wei2019single,li2020single,li2023two,dong2021location,hu2021trash,hu2023single,zhu2024revisiting,zhao2025reversible,zhong2024language,hong2024differ} improve performance by learning deep priors from data, but robust reflection removal in the wild remains challenging. More recently, diffusion models~\cite{zhou2025low,wang2025learning,lin2025harnessing,wu2024one,hong2024differ,hu2025dereflection} have shown strong potential for image restoration, which calls for appropriate guidance conditions to avoid undesired content drift and structural distortions. Overall, reflection removal faces an intrinsic trade-off, the tension between aggressive reflection suppression and faithful preservation of scene structure under large reflection variability. In practice, this trade-off manifests as three common issues on real mixtures, including incomplete reflection suppression, color inconsistency, and loss of fine details, as illustrated in~\cref{fig:intro_problem}.

To address this, we introduce an explicit spatial modulation mechanism that suppresses reflection strength spatially and preserves scene structure. We derive two complementary priors from the mixed image, a vision-language model (VLM)~\cite{li2025survey,wang2025internvl3,wei2025skywork,Qwen2.5-VL} based reflection intensity prior and a high-frequency prior by multi-scale residual decomposition. Examples of the priors $P_{\mathrm{int}}$ and $P_{\mathrm{hf}}$ are shown in~\cref{fig:intro_problem}. The intensity prior captures both region-level reflection analysis and global localization of reflection phenomena, thereby indicating where reflection removal should be stronger and to what extent. The high-frequency prior, in turn, provides a detailed sensitive signal that helps the model remain attentive to local structures during restoration. We integrate these priors into a spatial gate for selective modulation of the conditional residual injections.

Building on these guided priors, we propose \textbf{FUMO}, a dif\textbf{FU}sion model with prior \textbf{MO}dulation framework for SIRR, which adopts a coarse-to-fine design. In the first stage, a conditional diffusion model~\cite{rombach2022high,zhang2023adding} is guided by the proposed gate to aggressively remove reflections, driving the low frequency appearance toward the underlying transmission content. This stage delivers effective reflection removal, but strong suppression can still introduce geometric deviations or local inconsistencies. We therefore follow with a fine-grained refinement module, combined with the two priors, to correct distortions and restore coherent structures and details. Experiments demonstrate that our approach achieves strong performance under diverse scenes. 

The main contributions are summarized as follows: 
\begin{itemize}
    \item We introduce a dual-prior extraction pipeline for SIRR, supporting both reflection strength estimation and detail aware guidance. 
    \item We propose a prior-modulated diffusion framework that uses the two priors to guide conditional residual injection in a coarse-to-fine restoration process.
    \item Extensive experiments on benchmarks and in-the-wild images demonstrate competitive quantitative performance and improved visual quality.
\end{itemize}

\section{Related Work}
\label{sec:related_work}

%-------------------------------------------------------------------------
\subsection{Image Reflection Removal}

Due to the complex superposition of the transmission and reflection layers, recovering the transmission from a single blended image is a mathematically ill-posed problem. An intuitive solution is Multi-Image Reflection Removal (MIRR) \cite{Simon_2015_CVPR,yao2025polarfree,yang2016robust,szeliski2000layer,lei2020polarized,agrawal2005removing,han2017reflection}, which utilizes auxiliary information from multiple images, such as dynamic scenes~\cite{Simon_2015_CVPR}, focused/defocused images~\cite{szeliski2000layer}, or polarization imaging \cite{lei2020polarized,yao2025polarfree}. These methods mitigate reflection interference through comparative analysis and can effectively alleviate the ill-posed nature of the problem. However, their reliance on specific equipment or capture conditions limits their practical applicability, especially when processing existing images on the Internet or for mobile devices~\cite{hong2024differ}.

In contrast, Single-Image Reflection Removal (SIRR) ~\cite{Yang_2025_CVPR} aims to tackle this problem by exploring intrinsic priors from a single input. Some works relied on hand-crafted priors \cite{Arvanitopoulos_2017_CVPR,lei2021robust,zheng2021single,xue2015computational,levin2007user,shih2015reflection,zhong2024language}, such as edge annotations~\cite{levin2007user}, ghosting cues~\cite{shih2015reflection}, and sparsity assumptions~\cite{Arvanitopoulos_2017_CVPR}. With the development of deep learning, neural network-based methods \cite{wei2019single,li2020single,li2023two,dong2021location,hu2021trash,hu2023single,zhu2024revisiting,zhao2025reversible,zhong2024language,hong2024differ} have demonstrated superior performance and lower costs by autonomously learning implicit priors from the blended image. Dong~\etal~\cite{dong2021location} design a reflection detection module to regress a probabilistic reflection confidence map. YTMT~\cite{hu2021trash} enforces the predictions to communicate with each other. DSRNet~\cite{hu2023single} proposes DSFNet to initially extract transmission and reflection features. Zhu~\etal~\cite{zhu2024revisiting} introduces MaxRF to explicitly indicate virtual objects. RDNet~\cite{zhao2025reversible} employs a reversible encoder to secure valuable information. SIRR methods increasingly incorporate reflection localization and structure preserving designs to improve robustness in real scenes. Following this direction, we leverage explicit reflection strength priors together with detail oriented guidance to improve spatial adaptivity and reconstruction fidelity.

%-------------------------------------------------------------------------
\subsection{Diffusion Models}

In recent years, diffusion models~\cite{ho2020denoising,nichol2021improved,lu2025dpm,rombach2022high,podell2023sdxl,peebles2023scalable,bao2023all} have achieved significant progress. DDPM~\cite{ho2020denoising} establishes a powerful new paradigm and has sparked extensive follow up research. DDIM~\cite{nichol2021improved} further improves sampling efficiency by enabling a more flexible generation procedure. Latent diffusion models (LDMs)~\cite{rombach2022high,podell2023sdxl} conduct the diffusion process in a compact latent space and facilitate the integration of additional conditioning signals, which significantly broadens their applicability. In addition to U-Net denoisers, studies~\cite{peebles2023scalable,bao2023all} also explore diffusion transformer denoisers, further enriching architectural choices. These developments collectively strengthen the generative prior of diffusion models and provide flexible design choices for downstream tasks.

Generative models~\cite{dong2024ecmamba,zhou2024glare,liu2022griddehazenet+,liu2019griddehazenet,liu2021exploit,wu2024perception} have shown strong potential in image and video enhancement by leveraging learned priors for real-world degradations. With the rise of diffusion models as a powerful generative paradigm, diffusion-based methods have also been adapted for image enhancement and restoration, demonstrating powerful capabilities in tasks such as low-light enhancement~\cite{zhou2025low,jiang2024lightendiffusion,jiang2025learning}, dehazing~\cite{wang2025learning}, and super-resolution~\cite{lin2025harnessing,wu2024one}. In reflection removal, L-DiffER~\cite{hong2024differ} first introduces diffusion models to this task, where previous predictions are used as conditions to steer an iterative denoising process. DAI~\cite{hu2025dereflection} strengthens conditional injection with a ControlNet and further improves reconstruction with a refined decoder.  PolarFree~\cite{yao2025polarfree} leverages diffusion based modeling to produce reflection free imaging priors, which support subsequent reflection suppression. These advances suggest that diffusion models are promising for dereflection, and effective results often rely on appropriate conditioning and guidance to preserve faithful transmission content. Motivated by this insight, our framework adopts a diffusion backbone and injects priors through spatially adaptive modulation before fine-grained refinement.

\section{Methods}
The core of this work is to derive explicit guidance priors from the mixed image and use them to modulate diffusion-based reflection removal. 
The overall framework is illustrated in~\cref{fig:framework}.

In this section, we first introduce two complementary priors in~\cref{sec:prior_extract}. 
These priors form a spatial gate that modulates conditional residual injections in the coarse diffusion stage, followed by a lightweight refinement module for geometric consistency and detail coherence. The architecture and training objectives are described in~\cref{sec:framework} and~\cref{sec:P_training_strategy}.

\subsection{Dual Prior Extraction}
\label{sec:prior_extract}
\begin{figure}[t]
   \centering  
   \includegraphics[width=1\textwidth]{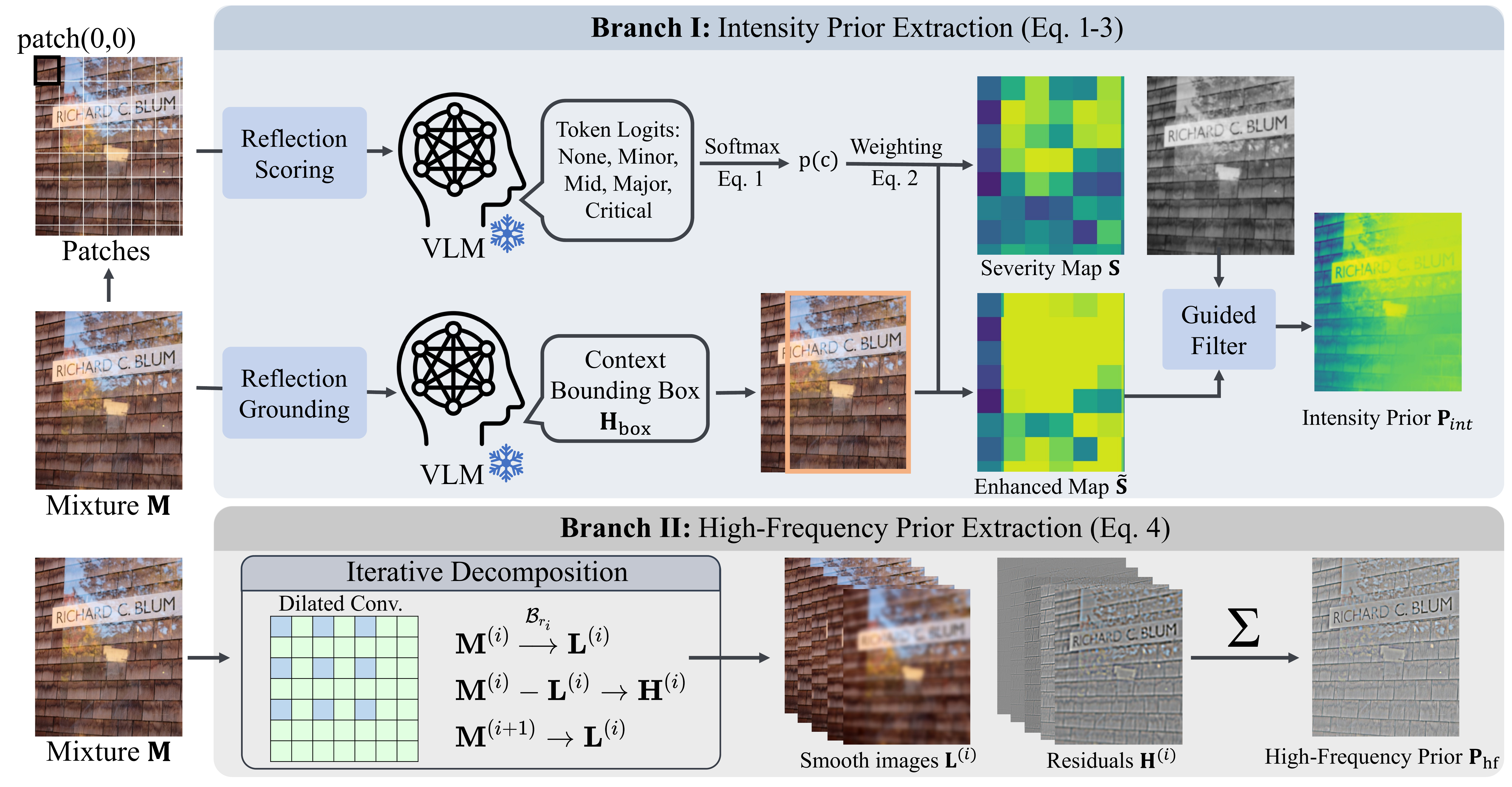}
   \caption{The pipeline of dual prior extraction. In branch I, the mixed image $\mathbf{M}$ is divided into patches, and the intensity prior $\mathbf{P}_{\mathrm{int}}$ is finally obtained through scoring and localization. In branch II, the mixed image $\mathbf{M}$ is iteratively decomposed and aggregated to yield the high-frequency prior $\mathbf{P}_{\mathrm{hf}}$.}
   \label{fig:priorframework}
\end{figure}

The mixed image alone usually lacks sufficient cues to robustly separate reflections from transmission structures. To alleviate this limitation, we extract two complementary priors, a VLM-derived intensity prior for reflection severity and a high-frequency prior for detail-sensitive responses. The overall prior extraction pipeline is shown in \cref{fig:priorframework}.

\subsubsection{Reflection Intensity Prior.} 

Vision Language Models (VLMs)~\cite{li2025survey,wang2025internvl3,wei2025skywork,Qwen2.5-VL,chen2023vlp} provide a practical interface for extracting semantic cues from a mixed image. Studies such as~\cite{wu2023q,liu2025step1x,chen2025multimodal,zhou2025low,wang2025hazeclip} suggest that VLM-derived signals, including intermediate reasoning outputs, can benefit image analysis and processing. Building on this capability, we use a frozen VLM to estimate a reflection intensity prior that indicates where reflection interference is stronger and to what extent. Given a mixed image $\mathbf{M}$, the reflection intensity prior $\mathbf{P}_{\mathrm{int}}\in[0,1]^{H\times W}$ is a pixel-level soft severity map that provides spatially grounded guidance for subsequent conditioning. 

We start from patch-level severity scoring to capture local reflection evidence at a manageable granularity. Inspired by Q-ALIGN~\cite{wu2023q}, we partition $\mathbf{M}$ into non-overlapping patches (with patch size $a$ chosen adaptively according to image resolution), and query a VLM to assess reflection severity using a fixed ordinal set $\mathcal{C}=\{\texttt{None},\texttt{Minor},\texttt{Mid},\texttt{Major},\texttt{Critical}\}$. Instead of relying on free-form text outputs, we exploit the model's next-token logits and restrict the probability mass to $\mathcal{C}$, yielding a more stable confidence distribution over the predefined levels:
\begin{equation}
p(c)=\frac{\exp(\ell(c)/\tau)}{\sum_{c'\in\mathcal{C}}\exp(\ell(c')/\tau)},\quad c\in\mathcal{C},
\label{eq:restricted_softmax}
\end{equation}
where $\ell(c)$ denotes the logit of the token corresponding to category $c$ and $\tau$ is a temperature. Then a continuous severity score is obtained through an ordinal expectation with weights $w(c)\in\{1,2,3,4,5\}$:
\begin{equation}
s=\sum_{c\in\mathcal{C}} w(c)\,p(c).
\label{eq:ordinal_expectation}
\end{equation}
We bring the patch scores back to the image plane to form a pixel-wise field $\mathbf{S}$ by assigning the same score to all pixels within each patch.

Patch-level scoring provides local estimates, but may overlook reflection regions that are visually prominent on the global scale. We further perform image-level analysis on the mixed image $\mathbf{M}$. Similarly to region localization in object detection, we prompt the VLM to identify areas dominated by reflections and return their bounding boxes. Then the corresponding regions in $\mathbf{S}$ are boosted with a multiplicative factor to obtain an enhanced map $\tilde{\mathbf{S}}$ (capped by a maximum value), which better captures large, coherent reflection patterns without altering the underlying patch scoring mechanism. On 100 real images with annotated reflection-dominant regions, this grounding step achieves 0.65 mIoU, serving as a coarse localization cue.

Subsequently, we densify $\tilde{\mathbf{S}}$ into a spatially coherent guidance map via an edge-aware guided filter~\cite{he2012guided}:
\begin{equation}
\mathbf{P}'_{\mathrm{int}}=\mathrm{GF}(\tilde{\mathbf{S}};\mathbf{G},r,\epsilon),
\label{eq:guided_filter}
\end{equation}
where the guide image $\mathbf{G}$ is obtained from $\mathbf{M}$ by converting to grayscale and applying a mild Gaussian pre-blur. This step removes block artifacts while encouraging intensity transitions to align with major image structures. Ultimately, we clamp the filtered response to a fixed range and linearly rescale it to $[0,1]$ to obtain the final intensity prior $\mathbf{P}_{\mathrm{int}}$, aligned with the input resolution.

\subsubsection{High-Frequency Prior.} 
The intensity prior provides spatial awareness of reflection strength. To complement it with local structural responses, we further extract a high-frequency prior from the same mixture input. Since it is extracted from the mixture $\mathbf{M}$, the prior contains high-frequency components from both the transmission and reflection layers. We therefore use it as a structure-sensitive guidance signal, which is combined with the intensity prior for spatial modulation.

Given the mixed image $\mathbf{M}$, we construct a multi-scale residual decomposition inspired by wavelet color correction~\cite{wang2024exploiting} to compute the high-frequency prior $\mathbf{P}_{\mathrm{hf}}$. Let $\mathcal{B}_{r}(\cdot)$ denote a channel-wise smoothing operator at scale $r$, implemented with dilated convolution~\cite{yu2015multi}. Starting from $\mathbf{M}^{(0)}=\mathbf{M}$, we iteratively compute a smoothed image and its residual at increasing scales:
$\mathbf{L}^{(i)}=\mathcal{B}_{r_i}(\mathbf{M}^{(i)})$,
$\mathbf{H}^{(i)}=\mathbf{M}^{(i)}-\mathbf{L}^{(i)}$,
and update $\mathbf{M}^{(i+1)}=\mathbf{L}^{(i)}$, where $r_i=2^i$ for $i=0,\ldots,L-1$.
The final high-frequency prior is obtained by accumulating residuals across scales:
\begin{equation}
\mathbf{P}_{\mathrm{hf}}=\sum_{i=0}^{L-1}\mathbf{H}^{(i)}.
\label{eq:hf_prior}
\end{equation}
By aggregating residual components from fine to coarse scales, $\mathbf{P}_{\mathrm{hf}}$ captures local structural responses while remaining simple and efficient to compute. We further clamp and linearly rescale the extracted response to $[0,1]$ at the input resolution for stable conditioning. Together with the intensity prior, it forms a complementary pair of explicit signals for the subsequent gated modulation design.

\subsection{Prior-modulated diffusion framework for SIRR}
 \label{sec:framework}
 \begin{figure}[t]
     \centering  
     \includegraphics[width=0.99\textwidth]{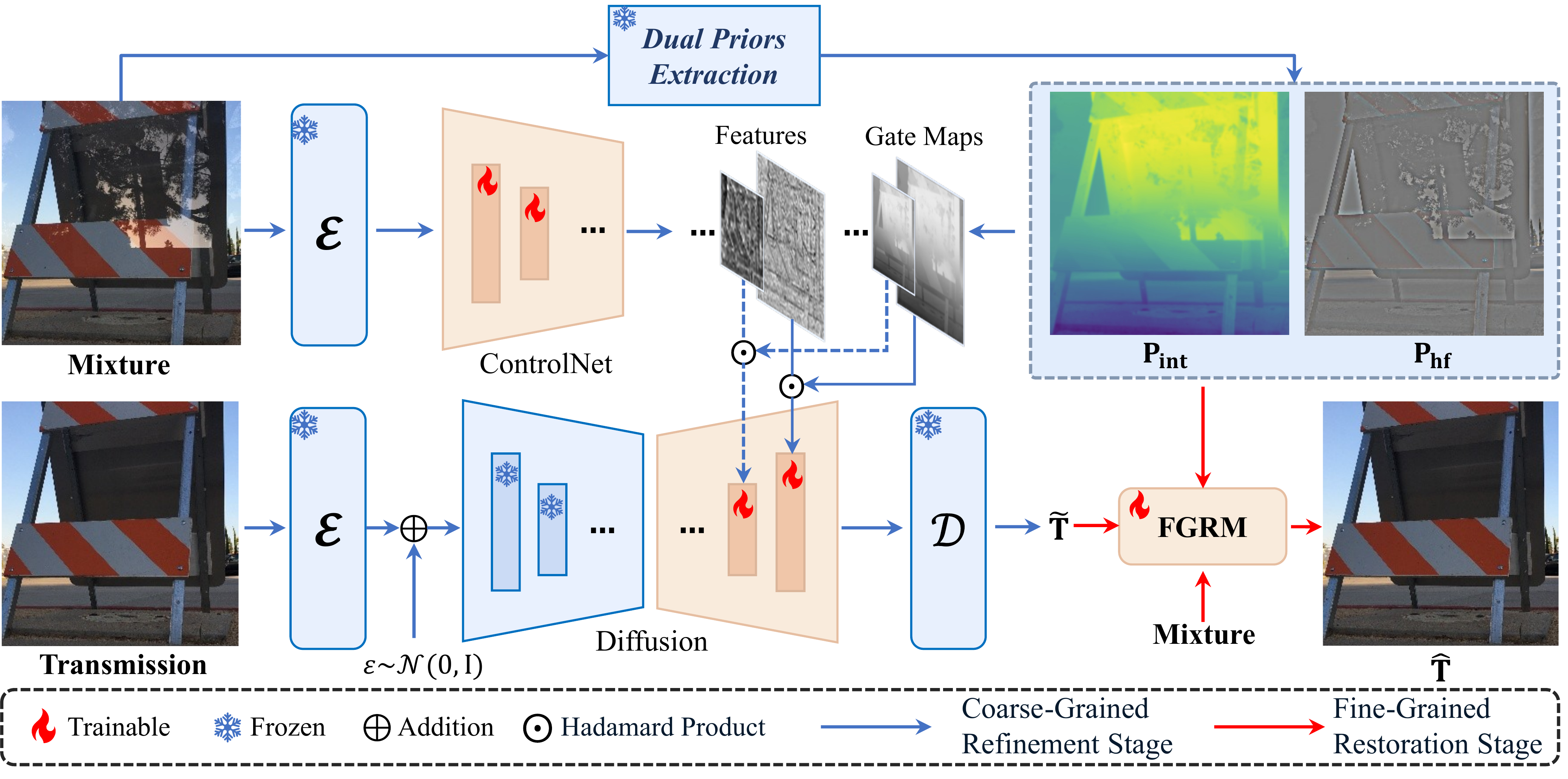}
     \caption{The framework of the proposed \textbf{FUMO} method. Given a mixed image $\mathbf{M}$, we obtain two priors $\mathbf{P}_{\mathrm{int}}$ and $\mathbf{P}_{\mathrm{hf}}$ through dual-prior extraction. A diffusion-based backbone performs conditional denoising, where the extracted features and gates are injected through element-wise fusion operations to guide multi-scale feature aggregation. The decoder produces a coarse restoration, which is further refined by the trainable fine-grained refinement module (FGRM).}
     \label{fig:framework}
 \end{figure}

As illustrated in \cref{fig:framework}, FUMO follows a coarse-to-fine prior-modulated diffusion design, consisting of a coarse restoration stage and a fine-grained refinement module.
The coarse stage adopts a ControlNet-conditioned~\cite{zhang2023adding} diffusion denoiser, where explicit priors are incorporated through a gated modulation mechanism for spatially adaptive conditioning. The refinement module further improves geometric consistency and restores fine structural details, producing the final transmission image.

\subsubsection{Coarse-Grained Restoration Diffusion Model.} 
The coarse stage targets strong reflection suppression to obtain an initial transmission estimate that is faithful in global appearance and low-frequency structure. We adopt a pretrained T2I diffusion model as the backbone, such as stable diffusion~\cite{rombach2022high}, which provides strong generative priors. The VAE encoder takes the mixed image $\mathbf{M}$ and transmission image $\mathbf{T}$ as input to generate latent tensors $z_\mathbf{M}$ and $z_\mathbf{T}$, respectively. ControlNet takes $z_\mathbf{M}$ as input to guide the SD model output. The diffusion denoiser ${\mu}_{\theta}$ receives noisy input $z_t=\alpha_t z_\mathbf{T}+ \sigma_t \epsilon$ (where $t$ denotes diffusion step, and $\alpha_t$ and $\sigma_t$ are determined by noise scheduler with $\epsilon \sim\mathcal N(0,\mathbf{I})$), and the control signal $c_m=f_\phi(\mathbf{M})$ from ControlNet $f_\phi$ to predict the latent $z_0 = z_\mathbf{T}$. Following~\cite{ye2024stablenormal,xu2024matters}, we adopt a one-step denoising formulation for the coarse restoration stage.

\subsubsection{Gated Modulation Module.} 

The intensity prior $\mathbf{P}_{\mathrm{int}}$ and the high-frequency prior $\mathbf{P}_{\mathrm{hf}}$ provide complementary guidance for spatially adaptive restoration. $\mathbf{P}_{\mathrm{int}}$ indicates where reflections are severe and thus require stronger conditional intervention, while $\mathbf{P}_{\mathrm{hf}}$ highlights structure-sensitive regions where residual modulation should be more attentive to local details. We combine them into a spatial gate that strengthens conditioning in regions with both severe reflection and rich local structure.

Specifically, we define the gate map as
\begin{equation}
\mathbf{g} \;=\; \mathbf{1} \;+\; \beta \, \mathbf{P}_{\mathrm{int}} \odot \mathbf{P}_{\mathrm{hf}},
\label{eq:gate}
\end{equation}
where $\odot$ denotes element-wise multiplication and $\beta$ controls the overall modulation strength. The additive identity term ensures that the mechanism reduces to standard conditioning when either guidance signal is weak, while the multiplicative interaction emphasizes their joint presence.

The gate is applied in the coarse stage to modulate the residual injection from the conditioning branch into the restoration network. Let $\mathbf{c}_m=\{\mathbf{c}_s\}$ denote the multi-scale residual tensors produced by ControlNet, where $s$ indexes the injection scale. For each scale, we resize $\mathbf{g}$ to match the spatial resolution of $\mathbf{c}_s$ and perform element-wise modulation to generate the gated modulation condition signal $\widetilde{\mathbf{c}}_s$:
\begin{equation}
\widetilde{\mathbf{c}}_s \;=\; \mathrm{clip}\!\left(\mathcal{I}_s(\mathbf{g}),\, 1,\, 1+\beta_{\max}\right)\odot \mathbf{c}_s,
\label{eq:gated_residual}
\end{equation}
where $\mathcal{I}_s(\cdot)$ denotes per-scale interpolation for alignment and $\mathrm{clip}(\cdot)$ clamps the gate to a bounded range for stable injection. In practice, $\beta$ is warmed up and then increased to $\beta_{\max}$ during training, so that the gated modulation is introduced progressively. We set $\beta_{\max}=0.25$ and use a warmup ratio of 0.1 empirically.

\subsubsection{Fine-Grained Refinement Module.} 

Although the coarse stage provides strong reflection suppression, its aggressive restoration may introduce geometric drift or local structural inconsistency, especially in regions where transmission structures are tightly entangled with reflections. To improve geometric coherence and recover fine details, we develop a fine-grained refinement module (FGRM) as a deterministic refiner operating in the image space.

The refinement module takes the coarse prediction together with the original mixture and the guidance signals as inputs. Let $\tilde{\mathbf{T}}$ denote the decoded transmission image obtained in the coarse stage. The refiner receives the channel-wise concatenation

\begin{equation}
\hat{\mathbf{T}} \;=\; R_{\phi}(\mathrm{concat}\!\left(\mathbf{M},\,\tilde{\mathbf{T}},\,\mathbf{P}_{\mathrm{hf}},\, \mathbf{P}_{\mathrm{int}} \right)),
\label{eq:refine_forward}
\end{equation}
where $R_{\phi}$ denotes the Fine-Grained Refinement Module. In practice, we adopt a U-Net but replace the standard activation with a lightweight channel gating operation that splits features into two halves and multiplies them element-wise, following the SimpleGate design in NAFNet~\cite{chen2022simple}.

\subsection{Training Strategy}
\label{sec:P_training_strategy}
The coarse-to-fine design is optimized separately with stage-specific objectives. The coarse stage focuses on one-step latent restoration under guided and gated modulation, while the refinement stage operates deterministically in the image space to improve geometric consistency and recover fine structures. 

\subsubsection{Coarse-Grained Restoration Stage.}
In the backward denoising process, conventional diffusion models predict noise at randomly sampled step $t$, and the corresponding multi-step training loss is formulated as:
\begin{equation}
\mathcal{L}_{\text{ldm}} \;=\; \bigl\|\epsilon - \mu_\theta(z_t,t,\widetilde{\mathbf{c}}_m)\bigr\|_2^2 .
\label{eq:loss_sd}
\end{equation}
where $\widetilde{\mathbf{c}}_m$ is the control signal from ControlNet and gated modulation.

Following~\cite{ye2024stablenormal,xu2024matters,hu2025dereflection}, we adopt one-step denoising for stable and efficient restoration. Instead of predicting noise, the network directly regresses a less perturbed latent. Specifically, for a target timestep $t$ (with a smaller $t$ indicating a cleaner latent), the model takes a maximally perturbed latent $z_N$ as input and predicts $z_t$ in a single forward pass:
\begin{equation}
\mathcal{L}_{\text{coarse}} \;=\; \bigl\|z_t - \mu_\theta(z_N,t,\widetilde{\mathbf{c}}_m)\bigr\|_2^2 .
\label{eq:loss_one_step}
\end{equation}
During training, $t$ is sampled uniformly and $N$ is set to 1000. At inference time, $t=0$ is set to obtain $\hat z_\mathbf{T}=\mu_\theta(z_N,0,\widetilde{\mathbf{c}}_m)$ with $z_N\sim\mathcal N(0,I)$, and $\hat z_\mathbf{T}$ is decoded to produce the restored transmission image.

In this training stage, only the ControlNet and the upsampling blocks of denoising U-Net are optimized. The VAE and the remaining parameters of U-Net are kept frozen to preserve the pretrained generative prior and stabilize optimization. 

\subsubsection{Fine-Grained Refinement Stage.}
The Fine-Grained Refinement Module refines the coarse prediction in the image space. Denoting the refined output by $\hat{\mathbf{T}}$ and the ground-truth transmission by $\mathbf{T}$, we employ a composite objective that balances pixel fidelity, perceptual similarity and geometric consistency. In this training stage, only the Fine-Grained Refinement Module is updated.

A pixel-level reconstruction loss constrains the global color and luminance:
\begin{equation}
\mathcal{L}_{\text{pix}} \;=\; \|\hat{\mathbf{T}} - \mathbf{T}\|_{1}.
\label{eq:loss_pix}
\end{equation}
Then we include a perceptual loss based on LPIPS~\cite{zhang2018unreasonable} to discourage over-smoothing and to better preserve visually salient structures:
\begin{equation}
\mathcal{L}_{\text{perc}} \;=\; \mathcal{L}_{\text{LPIPS}}(\hat{\mathbf{T}}, \mathbf{T})
=\sum_{i} w_i \| \phi_i(\hat{\mathbf{T}})-\phi_i(\mathbf{T})\|_2 ,
\label{eq:loss_perc}
\end{equation}
where $\phi_i(\cdot)$ denotes the feature map extracted from the $i$-th layer of AlexNet~\cite{krizhevsky2012imagenet}.
Finally, to explicitly promote edge alignment and mitigate local misalignment artifacts, we impose an edge-aware gradient loss and measure the discrepancy of both horizontal and vertical derivatives:
\begin{equation}
\mathcal{L}_{\text{grad}}=\lVert \nabla_x \hat{\mathbf{T}}-\nabla_x \mathbf{T}\rVert_1+\lVert \nabla_y \hat{\mathbf{T}}-\nabla_y \mathbf{T}\rVert_1
\label{eq:loss_grad}
\end{equation}
where $\nabla(\cdot)$ denotes image gradients computed by fixed Sobel filters.

Gathering all the loss terms yields the refinement objective as:
\begin{equation}
\mathcal{L}_{\text{refine}} \;=\; 
\lambda_{\text{pix}}\,\mathcal{L}_{\text{pix}} \;+\;
\lambda_{\text{perc}}\,\mathcal{L}_{\text{perc}} \;+\;
\lambda_{\text{grad}}\,\mathcal{L}_{\text{grad}},
\label{eq:loss_refine}
\end{equation}
where the weights $\lambda_{\text{pix}} = 0.5$, $\lambda_{\text{perc}} = 0.25$, and $\lambda_{\text{grad}} = 0.25$ are set empirically.

\section{Experiments}

\subsection{Implementation Details}
We implement the proposed method in PyTorch~\cite{paszke2019pytorch}. All experiments are conducted on 4 NVIDIA GeForce RTX 4090 GPUs with a batch size of 1. The two stages are trained separately, with 100k steps and 10k steps respectively. In the coarse stage, we initialize the U-Net and ControlNet with pretrained weights of Stable Diffusion V2.1~\cite{rombach2022high}, and update the weights using AdamW~\cite{kingma2014adam} with a learning rate of $5 \times 10^{-5}$. In the refinement stage, we freeze the coarse restoration part and train the fine-grained refinement module using AdamW with a learning rate of $1 \times 10^{-4}$. At $1$K resolution on one RTX 4090, prior extraction takes $3.40$s, and the restoration network takes $0.63$s with $1.72$B parameters, $1512.96$G FLOPs, and $4.96$GB peak memory.

\subsubsection{Datasets}
Training uses a combination of real paired data and synthetic data. The real subset contains 89 image pairs from Real~\cite{zhang2018single}, 200 pairs from Nature~\cite{li2020single}, 1230 pairs from RR4k~\cite{chen2024real}, 6600 pairs from RRW~\cite{zhu2024revisiting} and 23303 pairs from DRR~\cite{hu2025dereflection}. The remaining image pairs from Real and Nature are used for evaluation. For synthetic data, we generate 16120 training pairs by blending images sampled independently from the COCO~\cite{lin2014microsoft} dataset, using formulation from DSRNet~\cite{hu2023single} $M \;=\; \gamma_1 T +\gamma_2 R -\gamma_1 \gamma_2 T \odot R$. Following~\cite{zhao2025reversible}, we sample $\gamma_1$ and $\gamma_2$ for 3 channels individually. During training, all images are resized and randomly cropped to a resolution of 768, with random flipping and color jitter applied for augmentation.

\subsection{Comparison with State-of-the-Art Methods}
\label{sec:compare}

\begin{figure}[t]
    \centering
    \includegraphics[width=\linewidth]{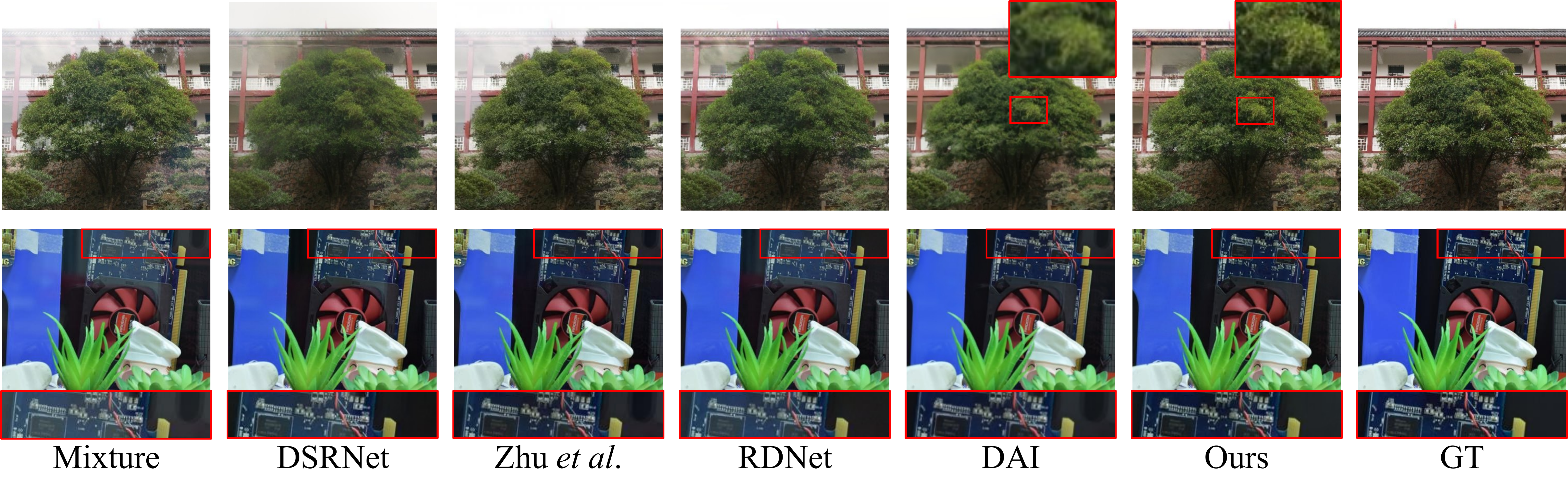}
    \caption{Qualitative comparisons on representative examples from the three benchmarks. The red rectangles highlight key regions for comparison. }
    \label{fig:quality_bench}
\end{figure}
\begin{figure}[t]    
    \centering
    \includegraphics[width=\linewidth]{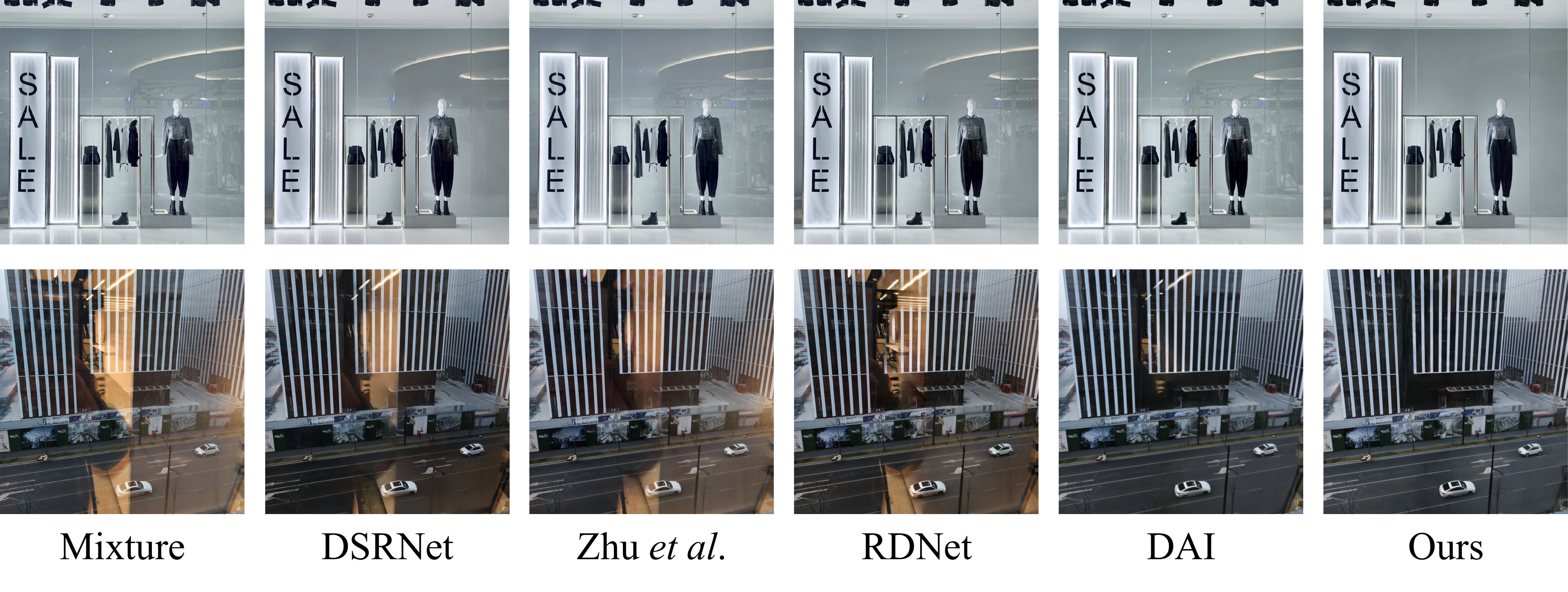}
    \caption{Qualitative comparisons on challenging real-world mixed images. }
    \label{fig:quality_wild}
\end{figure}

We evaluate reflection removal on three widely used real benchmarks, including Nature~\cite{li2020single}, Real~\cite{zhang2018single}, and SIR2~\cite{wan2017benchmarking}, following their standard test splits and evaluation protocols. To assess robustness under unconstrained imaging conditions, we further report qualitative results on a small set of in-the-wild mixed images collected from the Internet, which exhibit diverse reflection patterns and challenging scene content.
We compare our method with representative single image reflection removal approaches, including IBCLN~\cite{li2020single}, Dong~\etal~\cite{dong2021location}, YTMT~\cite{hu2021trash}, DSRNet~\cite{hu2023single}, Zhu~\etal~\cite{zhu2024revisiting}, RDNet~\cite{zhao2025reversible} and DAI~\cite{hu2025dereflection}. For methods with publicly available training code, we follow the official implementations and finetune them on our training data to reduce the impact of mismatched data distributions. All competing methods are evaluated using the same test images, preprocessing pipeline, and metrics.

\subsubsection{Quantitative Comparison.}
In quantitative evaluation, we employ PSNR~\cite{huynh2008scope} and SSIM~\cite{wang2003multiscale} to measure fidelity with respect to the ground truth transmission, together with LPIPS~\cite{zhang2018unreasonable} for perceptual similarity. We additionally report CLIPIQA~\cite{wang2023exploring} and MUSIQ~\cite{ke2021musiq} as complementary no-reference quality indicators. Quantitative results are summarized in~\cref{tab:quant_main}. Across the three benchmarks, our method remains competitive on distortion-oriented metrics (PSNR/SSIM) and shows clear advantages on perceptual metrics (LPIPS/CLIPIQA/MUSIQ), indicating that the proposed guidance and gated modulation improve visual quality while better preserving transmission structures under complex reflections.

\begin{table*}[t]
\centering
\setlength{\tabcolsep}{4.2pt}    
\renewcommand{\arraystretch}{1.08} 
\caption{Quantitative comparisons on three reflection removal benchmarks (Nature, Real, and SIR$^2$).
For each evaluation metric, the arrow $\uparrow$ ($\downarrow$) indicates that larger (smaller) values are better. The best results are highlighted in \textbf{bold},
and the second-best results are \underline{underlined}.}
\label{tab:quant_main}
\resizebox{\textwidth}{!}{%
\begin{tabular}{cccccccccc}
\toprule
\multirow{2}{*}{\makecell{Dataset\\(size)}} &
\multirow{2}{*}{Metric} &
\multicolumn{8}{c}{Method} \\
\cmidrule(lr){3-10}
& & IBCLN & Dong~\etal & YTMT & DSRNet & Zhu~\etal & RDNet & DAI & Ours \\
\midrule

\multirow{5}{*}{\makecell{Nature\\(20)}} 
& PSNR$\uparrow$ & 23.77 & 23.33 & 20.76 & 24.58 & 25.68 & 25.74 & \second{26.81} & \best{26.93} \\
& SSIM$\uparrow$ & 0.786 & 0.812 & 0.768 & 0.817 & 0.826 & \second{0.828} & \best{0.840} & \best{0.840} \\
& LPIPS$\downarrow$ & 0.145 & 0.117 & 0.183 & 0.120 & \second{0.103} & 0.109 & 0.203 & \best{0.088} \\
& CLIPIQA$\uparrow$  & 0.408 & 0.419 & 0.400 & \best{0.440} & 0.432 & 0.431 & 0.362 & \second{0.439} \\
& MUSIQ$\uparrow$   & 58.83  & 59.56  & 59.63  & \second{62.09}  & 61.37  & 60.26  & 54.70  & \best{62.87} \\
\midrule

\multirow{5}{*}{\makecell{Real\\(20)}} 
& PSNR$\uparrow$ & 21.55 & 22.34 & 22.87 & 23.37 & 21.85 & 24.81 & \second{25.21} & \best{25.95} \\
& SSIM$\uparrow$ & 0.767 & 0.811 & 0.808 & 0.801 & 0.781 & \second{0.838} & \second{0.838} & \best{0.852} \\
& LPIPS$\downarrow$ & 0.210 & 0.150 & 0.157 & 0.157 & 0.183 & \second{0.118} & 0.150 & \best{0.097} \\
& CLIPIQA$\uparrow$  & 0.444 & 0.441 & 0.464 & 0.483 & 0.453 & \best{0.538} & 0.414 & \second{0.509} \\
& MUSIQ$\uparrow$   & 58.08  & 57.55  & 58.72  & 60.44  & 58.46  & \second{61.42}  & 58.20  & \best{62.12} \\
\midrule

\multirow{5}{*}{\makecell{SIR$^2$\\(500)}} 
& PSNR$\uparrow$ & 23.89 & 24.29 & 23.73 & 25.51 & 25.37 & 26.46 & \best{27.35} & \second{27.22} \\
& SSIM$\uparrow$ & 0.886 & 0.902 & 0.891 & 0.913 & 0.904 & 0.921 & \second{0.923} & \best{0.925} \\
& LPIPS$\downarrow$ & 0.127 & 0.099 & 0.119 & 0.094 & 0.112 & \second{0.080} & 0.093 & \best{0.067} \\
& CLIPIQA$\uparrow$  & 0.384 & 0.394 & 0.385 & 0.405 & \second{0.408} & 0.396 & 0.365 & \best{0.419} \\
& MUSIQ$\uparrow$   & 58.16  & 57.55  & 58.09  & \second{58.82}  & 58.11  & 58.53  & 54.40  & \best{59.85} \\
\midrule

\multirow{5}{*}{\makecell{Average\\(540)}} 
& PSNR$\uparrow$ & 23.79 & 24.17 & 23.57 & 25.39 & 25.24 & 26.36 & \best{27.24} & \second{27.15} \\
& SSIM$\uparrow$ & 0.877 & 0.895 & 0.882 & 0.905 & 0.896 & 0.914 & \second{0.916} & \best{0.918} \\
& LPIPS$\downarrow$ & 0.131 & 0.102 & 0.123 & 0.098 & 0.114 & \second{0.083} & 0.100 & \best{0.069} \\
& CLIPIQA$\uparrow$  & 0.387 & 0.397 & 0.388 & 0.410 & \second{0.411} & 0.403 & 0.367 & \best{0.424} \\
& MUSIQ$\uparrow$   & 58.19  & 57.63  & 58.18  & \second{59.02}  & 58.26  & 58.71  & 54.57  & \best{59.88} \\
\bottomrule
\end{tabular}}
\end{table*}

\subsubsection{Qualitative Comparison.}

\cref{fig:quality_bench} shows qualitative comparisons on representative examples from the three benchmarks.
In addition, \cref{fig:quality_wild} presents results on extra real-world mixed images, including photos captured by us and images collected from the Internet, to further assess robustness under diverse reflection conditions.
The results demonstrate that previous methods often struggle under complex reflections, leading to residual reflection traces, local over-smoothing, or color inconsistencies. In contrast, our results better retain geometric coherence and fine structures, particularly around edges and textured regions, while removing reflection patterns that are entangled with transmission content.

\subsection{Ablation Study}

\subsubsection{Ablation on Gated Modulation.}
\begin{figure}[t]
    \centering
    \includegraphics[width=\linewidth]{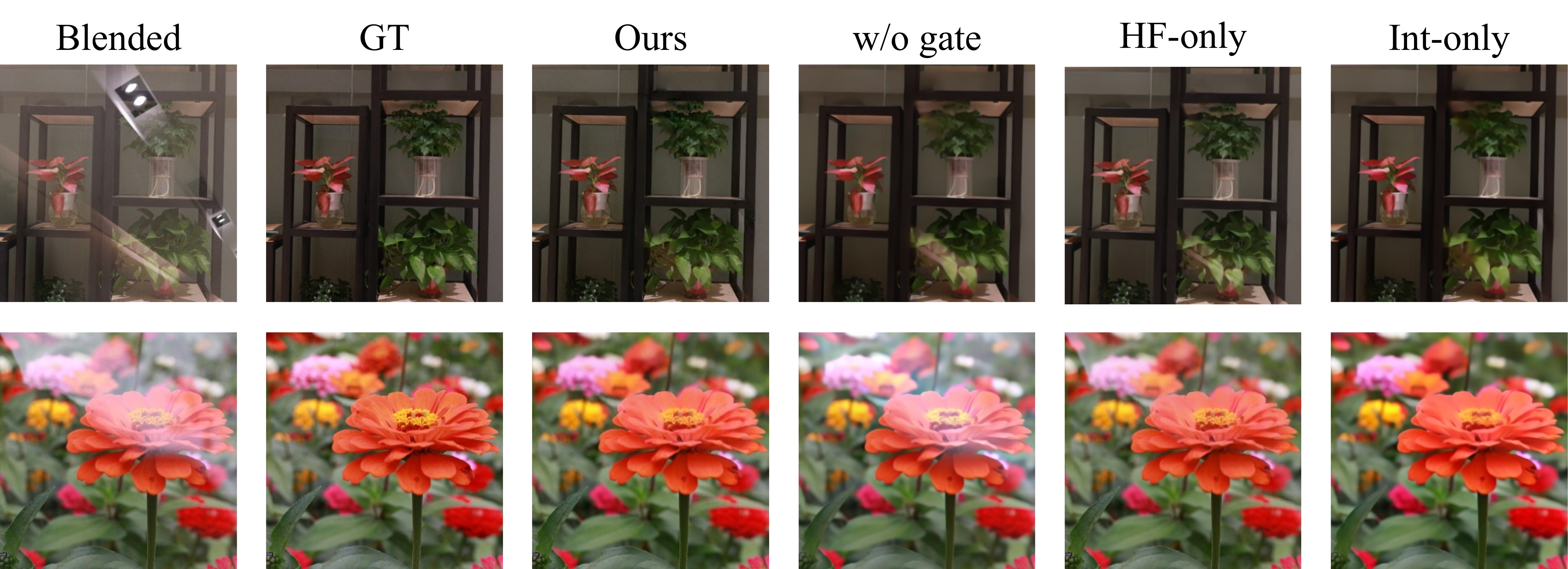}
    \caption{Qualitative ablation results on gated modulation. }
    \label{fig:ablation_gating}
\end{figure}

This part isolates the role of gated modulation in the coarse stage, where conditional residual injections are spatially modulated before entering the diffusion denoiser. All variants share the same training methods, and the only change lies in how the gate is formed from the guidance signals.
Specifically, we compare: (i) \emph{w/o gate}, where the injection tensors are passed without modulation ($g\equiv 1$); (ii) \emph{intensity only}, where the gate is driven solely by the intensity prior ($g=1+\beta P_{\mathrm{int}}$); (iii) \emph{high-frequency only}, where the gate is driven solely by the high-frequency prior ($g=1+\beta P_{\mathrm{hf}}$); and (iv) \emph{full}, where the gate combines both signals as in~\cref{eq:gate}.
We additionally evaluate a \emph{concat} variant that directly concatenates $\mathbf{P}_{\mathrm{int}}$ and $\mathbf{P}_{\mathrm{hf}}$ with the ControlNet condition instead of modulating residual injections.
\cref{tab:ablation_quant} reports quantitative results on the same benchmarks and metrics as in~\cref{sec:compare}, and \cref{fig:ablation_gating} provides representative visual comparisons.

Across datasets, removing gated modulation degrades perceptual quality most noticeably and the outputs exhibit more residual reflection artifacts and more obvious restoration traces in \cref{fig:ablation_gating}.
Using a single guidance signal partially alleviates these artifacts, but the two priors emphasize different failure modes. The intensity driven gate primarily improves suppression in high severity regions, whereas the high-frequency driven gate improves responses around edges and textured regions.
Direct concatenation improves over the no-gate baseline, but remains weaker than gated modulation, indicating that the priors are more effective when used to modulate residual injections. Combining $P_{\mathrm{int}}$ and $P_{\mathrm{hf}}$ in the gate consistently gives the most favorable perceptual scores among the variants and produces cleaner reflection removal with more faithful transmission structures, supporting the design choice of modulating coarse stage injections with a severity aware and detail sensitive signal.

\begin{table*}[t]
\centering
\setlength{\tabcolsep}{6.0pt}    
\renewcommand{\arraystretch}{1.10} 
\caption{Quantitative ablation results on gated modulation and refinement module.
For each evaluation metric, the arrow $\uparrow$ ($\downarrow$) indicates that larger (smaller) values are better. The best results are highlighted in \textbf{bold}. HF-only refers to \emph{high-frequency only} variant}
\label{tab:ablation_quant}
\resizebox{\textwidth}{!}{%
\begin{tabular}{cccccccc}
\toprule
\multirow{2}{*}{Metric} &
\multicolumn{4}{c}{Abl.\ on gated modulation} &
\multicolumn{2}{c}{Abl.\ on refinement module} \\
\cmidrule(lr){2-5}\cmidrule(lr){6-7}
& W/o gate & Concat & Intensity-only & HF-only & Coarse-only & Fine-decoder & Ours \\
\midrule
PSNR$\uparrow$  & 26.55 & 26.87 & 26.94 & 26.78 & 26.27 & 26.75 & \textbf{27.15} \\
SSIM$\uparrow$  & 0.897 & 0.906 & 0.905 & 0.906 & 0.865 & 0.891 & \textbf{0.918} \\
LPIPS$\downarrow$ & 0.076 & 0.072 & 0.073 & 0.071 & 0.133 & 0.084 & \textbf{0.069} \\
CLIPIQA$\uparrow$  & 0.390 & 0.406 & 0.403 & 0.393 & 0.385 & 0.398 & \textbf{0.424} \\   
MUSIQ$\uparrow$   & 57.13 & 58.55 & 58.40 & 58.08 & 57.56 & 58.43 & \textbf{59.88} \\
\bottomrule
\end{tabular}%
}
\end{table*}

\subsubsection{Ablation on Refinement Module.}

This experiment examines the contribution of the Fine-Grained Refinement by comparing the full pipeline with two variants: (i) the \emph{coarse-only} variant and (ii) the \emph{fine-decoder} variant. While the \emph{coarse-only} variant directly decodes the Coarse-Grained Restoration output as the final prediction, the \emph{fine-decoder} variant utilizes a fine-tuned decoder following the training protocol in DAI~\cite{hu2025dereflection}. 
All settings use the same coarse model. 

\begin{figure}[t]
    \centering
    \includegraphics[width=\linewidth]{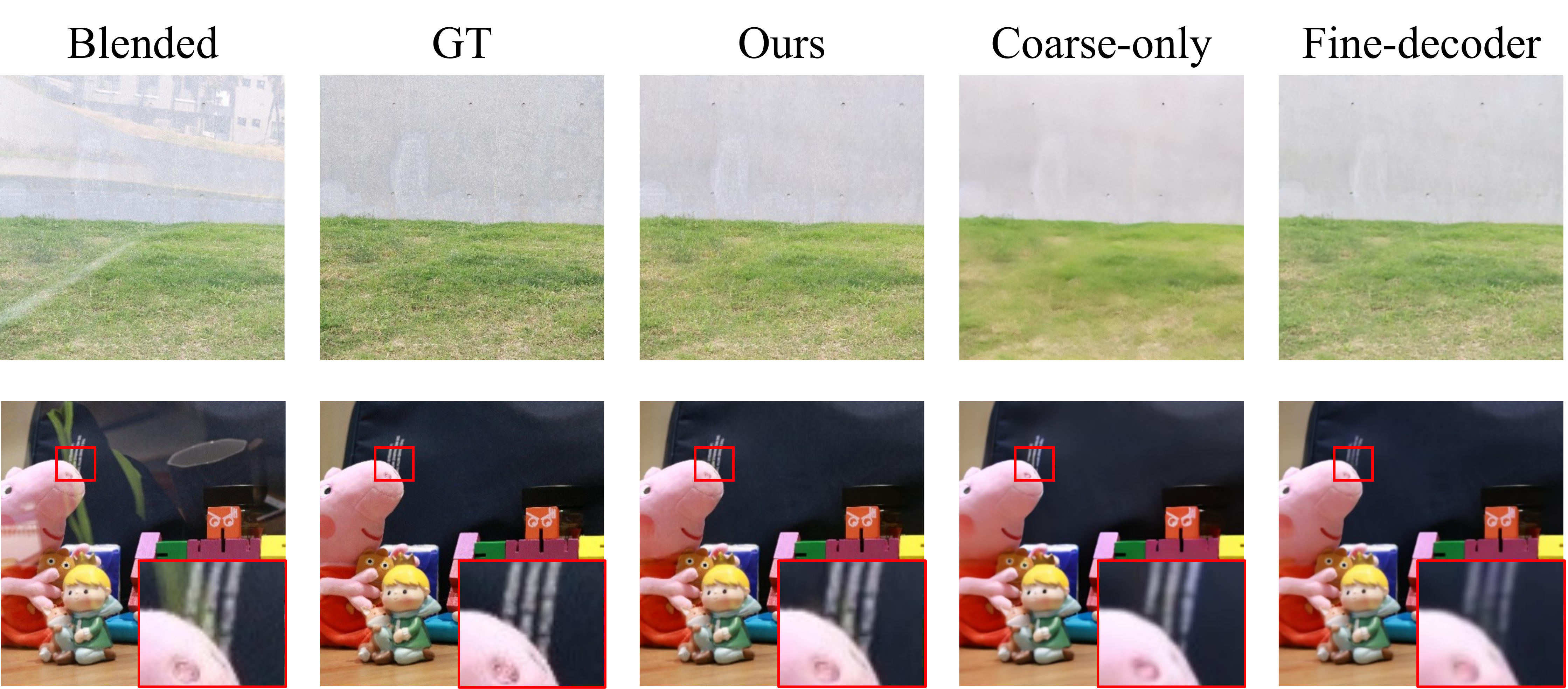}
    \caption{Qualitative ablation results on refinement module. }
    \label{fig:ablation_refine}
\end{figure}

As summarized in \cref{tab:ablation_quant} and illustrated in \cref{fig:ablation_refine}, the \emph{coarse-only} variant is already effective at suppressing dominant reflections, confirming the capability of the one-step coarse restoration. However, its outputs often exhibit noticeably reduced sharpness and visible restoration traces, particularly around edges and textured regions where reflection patterns intertwine with transmission structures. With a fine-tuned decoder, \emph{fine-decoder} variant shows partial improvements in artifact removal and detail enhancement, but falls short of fully addressing the underlying problem.  Introducing the refinement module largely improves visual clarity and structural coherence. Residual blur and local artifacts are substantially reduced. These observations support the role of the refinement network as a deterministic correction step that consolidates the coarse restoration into a sharper and more geometrically consistent transmission estimate.

\section{Conclusion}
This paper proposed a diffusion framework for single image reflection removal that improves spatial adaptivity and structural faithfulness under varying reflections. The method derives two guidance priors from the mixed image and uses them to guide restoration in a complementary manner. The coarse stage applies gated modulation to strengthen reflection suppression where it is needed and to better preserve transmission structures in detail sensitive regions. The refinement module improves geometric consistency and visual clarity and reduces blur and restoration traces. Experiments on standard benchmarks and additional real-world mixtures show competitive performance and clear improvements on perceptual quality measures.

\subsubsection{Limitations.}
The method can be ambiguous when reflection and transmission content overlap with similar intensity and similar structures. In such cases, it is difficult to determine which content should be preserved. User-provided spatial guidance such as coarse masks or sparse edge annotations is a practical direction to reduce this ambiguity. In addition, the VLM-derived prior extraction introduces extra inference overhead compared with purely feed-forward restoration networks. It can be mitigated by replacing prior extraction with a lightweight predictor.

%\clearpage\mbox{}Page \thepage\ of the manuscript. This is the last page.
%\par\vfill\par
%Now we have reached the maximum length of an ECCV \ECCVyear{} submission (excluding references).
%References should start immediately after the main text, but can continue past p.\ 14 if needed.
% \clearpage  % TODO REVIEW/FINAL: This \clearpage needs to be removed from both review and camera-ready versions.

\section*{Acknowledgements}
% % % % Please insert your acknowledgments here.
% We thank anonymous reviewers for their insightful comments and constructive suggestions that greatly improved this paper.
The work was supported in part by the National Natural Science Foundation of China under Grant 62301310 and 62572317.

\bibliographystyle{splncs04}
\bibliography{main}
\end{document}